\documentclass[11pt,a4paper]{article}
\usepackage[hyperref]{acl2018}
\usepackage{times}
\usepackage{latexsym}
\usepackage[pdftex]{graphicx}
\usepackage{url}
\usepackage{caption}
\usepackage{comment}
\usepackage{subfigure}
\usepackage{here}

\usepackage{multirow}

\aclfinalcopy % Uncomment this line for the final submission

\title{Multi-Source Neural Machine Translation with Missing Data}

\author{Yuta Nishimura\(^{1}\) , Katsuhito Sudoh\(^{1}\) , Graham Neubig\(^{2,1}\), Satoshi Nakamura\(^{1}\) \\
     \(^{1}\)Nara Institute of Science and Technology, 8916-5 Takayama-cho, Ikoma, Nara 630-0192, Japan\\
     \(^{2}\)Carnegie Mellon University, 5000 Forbes Avenue, Pittsburgh, PA 15213, USA\\
    {\tt \{nishimura.yuta.nn9, sudoh, s-nakamura\}@is.naist.jp} \\  
    {\tt gneubig@cs.cmu.edu} \\  
   }

\date{}

\begin{document}
\maketitle
\begin{abstract}
  Multi-source translation is an approach to exploit multiple inputs (e.g. in two different languages) to increase translation accuracy.
  In this paper, we 
  examine approaches for multi-source neural machine translation (NMT) using an \textit{incomplete} multilingual corpus
  in which some translations are missing.
  In practice, many multilingual corpora are not complete due to the difficulty to provide translations in \emph{all} of the relevant languages (for example, in TED talks, most English talks only have subtitles for a small portion of the languages that TED supports).
  Existing studies on multi-source translation did not explicitly handle such situations.
  This study focuses on the use of incomplete multilingual corpora in multi-encoder NMT and mixture of NMT experts and examines a very simple implementation where missing source translations are replaced by a special symbol $<$NULL$>$.
  These methods allow us to use incomplete corpora both at training time and test time.
  In experiments with real incomplete multilingual corpora of TED Talks, the multi-source NMT with the $<$NULL$>$ tokens
  achieved higher translation accuracies measured by BLEU than those by any one-to-one NMT systems.
\end{abstract}

\begin{figure}[t]
	\subfigure[A standard bilingual corpus]{
      \includegraphics[width=7.5cm]{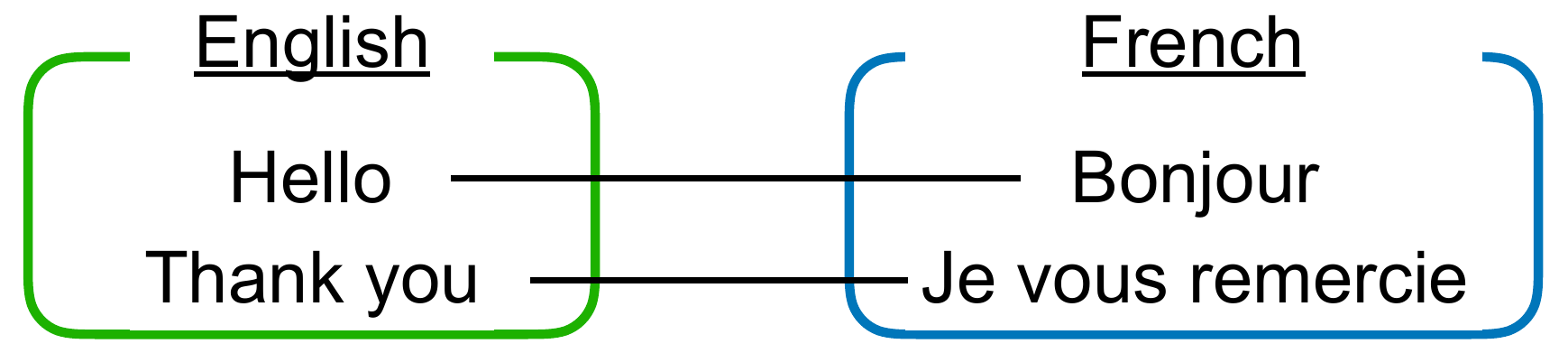}}
	\subfigure[A complete multi-source corpus]{
      \includegraphics[width=7.5cm]{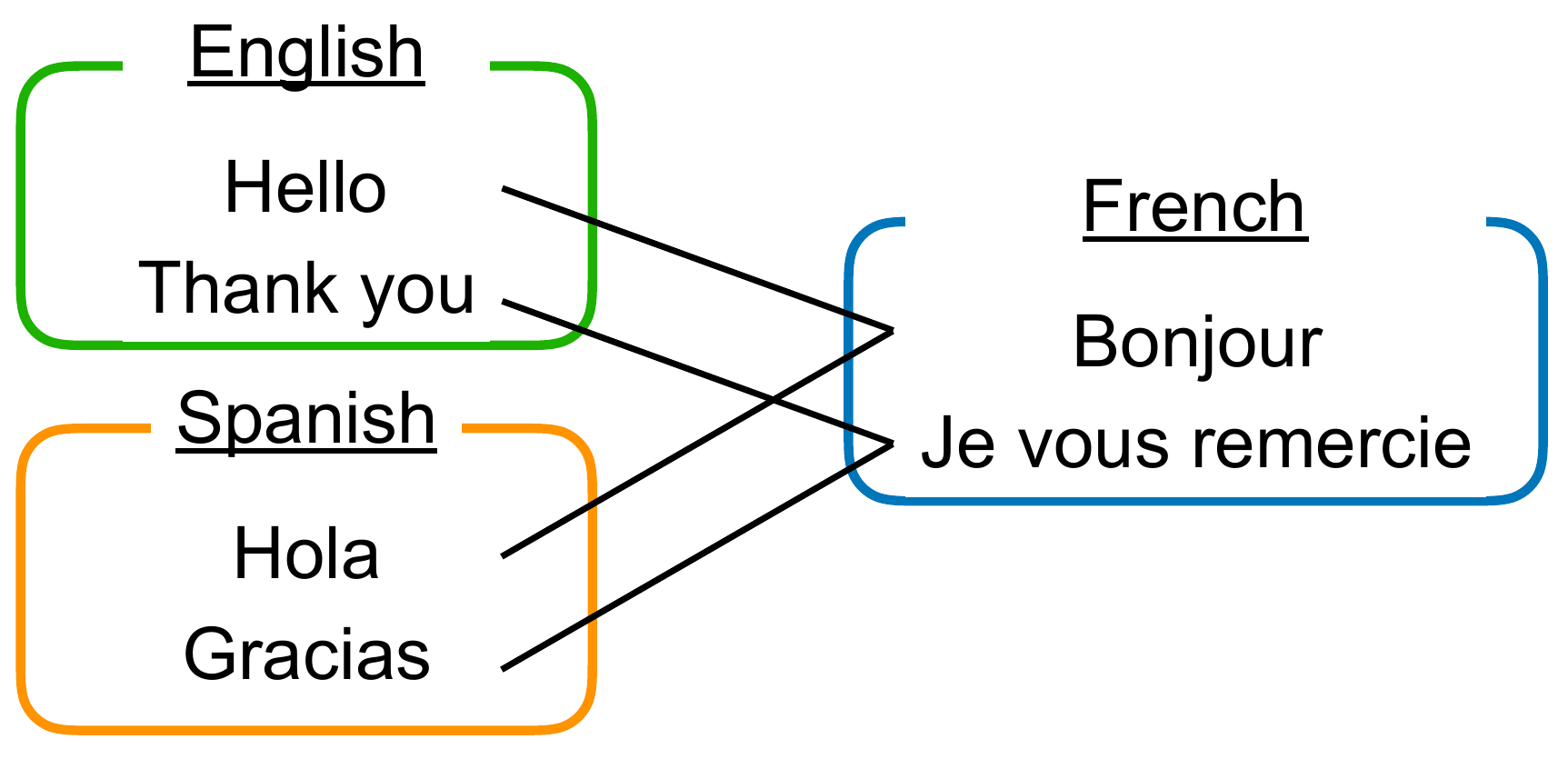}}
	\subfigure[An incomplete multi-source corpus with missing data]{
      \includegraphics[width=7.5cm]{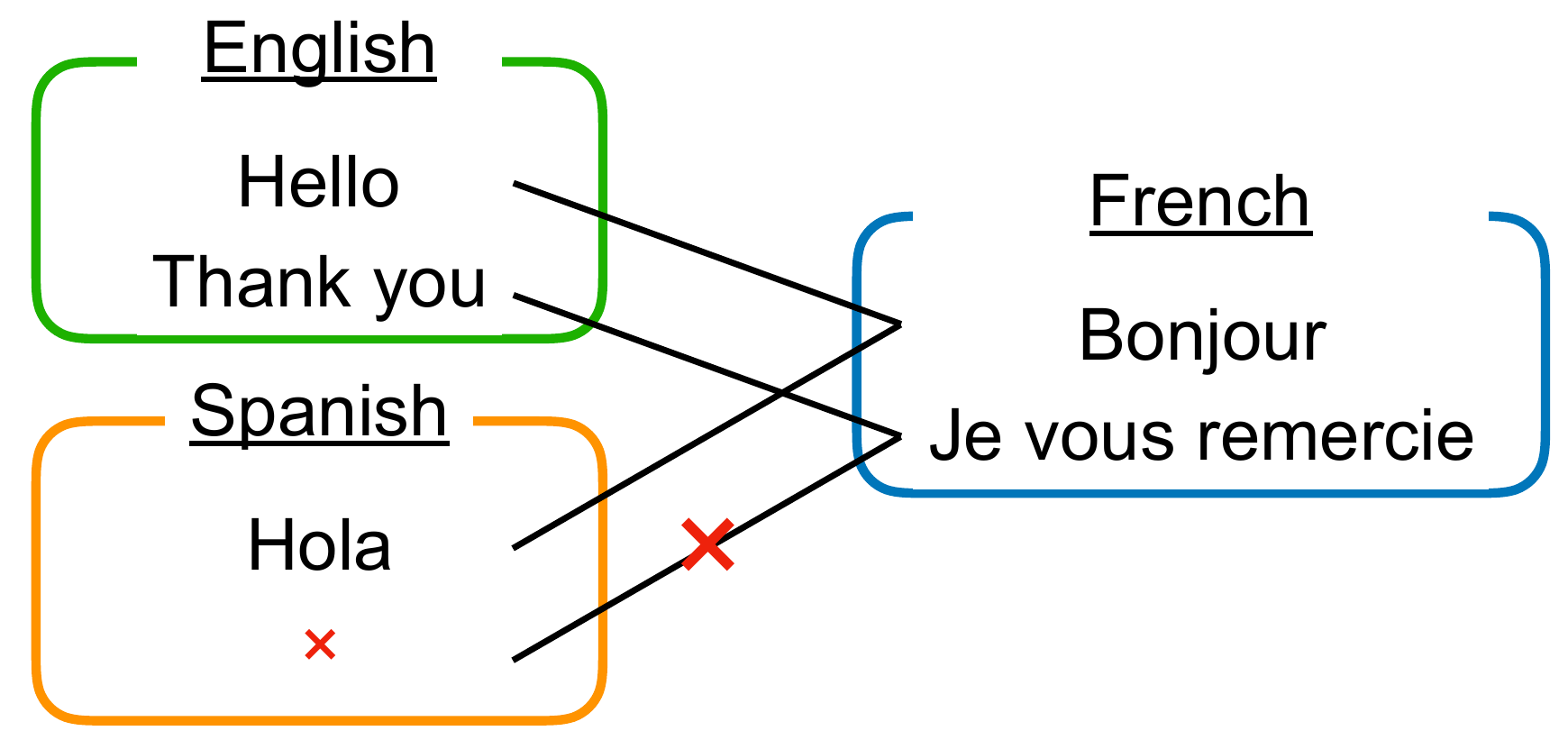}}
    \caption{Example of type of corpora.} 
    \label{fig:incomp_corpora}
\end{figure}

\section{Introduction}

In general, machine translation systems translate from one source language to a target language.
For example, we may translate a document or speech that was written in English to a new language such as French.
However, in many real translation scenarios, there are cases where there are multiple languages involved in the translation process.
For example, we may have an original document in English, that we want to translate into several languages such as French, Spanish, and Portuguese.
Some examples of these scenarios are the creation of video captions for talks \cite{Cettolo-etc:2012:eamt} or Movies \cite{Tiedemann:RANLP5}, or translation of official documents into all the languages of a governing body, such as the European parliament \cite{koehn2005epc} or UN \cite{ZIEMSKI16.1195}.
In these cases, we are very often faced with a situation where we \emph{already} have good, manually curated translations in a number of languages, and we'd like to generate translations in the remaining languages for which we do not yet have translations.

In this work, we focus on this sort of multilingual scenario using multi-source translation
\cite{Och2001,zoph-knight:2016:N16-1,garmash-monz:2016:COLING}.
Multi-source translation takes in multiple inputs, and references all of them when deciding which sentence to output.
Specifically, in the context of neural machine translation (NMT), there are several methods proposed to do so.
For example, \newcite{zoph-knight:2016:N16-1} propose a method where multiple sentences are each encoded separately, then all referenced during the decoding process (the ``multi-encoder'' method).
In addition, \newcite{garmash-monz:2016:COLING} propose a method where NMT systems over multiple inputs are ensembled together to make a final prediction (the ``mixture-of-NMT-experts'' method).

However, this paradigm assumes that we have data in \emph{all} of the languages that go into our multi-source system. For example, if we decide that English and Spanish are our input languages and that we would like to translate into French, we are limited to training and testing only on data that contains all of the source languages.
However, it is unusual that translations in all of these languages are provided-- there will be many sentences where we have only one of the sources.
In this work, we consider methods for multi-source NMT with missing data, such situations using an \textit{incomplete} multilingual corpus in which some translations are missing, as shown in Figure~\ref{fig:incomp_corpora}.
This incomplete multilingual scenario is useful in practice, such as when creating translations for incomplete multilingual corpora such as subtitles for TED Talks.

In this paper, we examine a simple implementation of multi-source NMT using such an incomplete multilingual corpus
that 
uses a special symbol $<$NULL$>$ to represent the missing sentences.
This can be used with any existing multi-source NMT implementations without no special modifications.
Experimental results with real incomplete multilingual corpora of TED Talks show that
it is effective in allowing for multi-source NMT in situations where full multilingual corpora are not available, resulting in BLEU score gains of up to 2 points compared to standard bi-lingual NMT.

\section{Multi-Source NMT}

At the present, there are two major approaches to multi-source NMT: multi-encoder NMT \cite{zoph-knight:2016:N16-1} and mixture of NMT experts \cite{garmash-monz:2016:COLING}.
We first review them in this section.

\subsection{Multi-Encoder NMT}
\begin{figure}[t]
	\includegraphics[width=7.5cm]{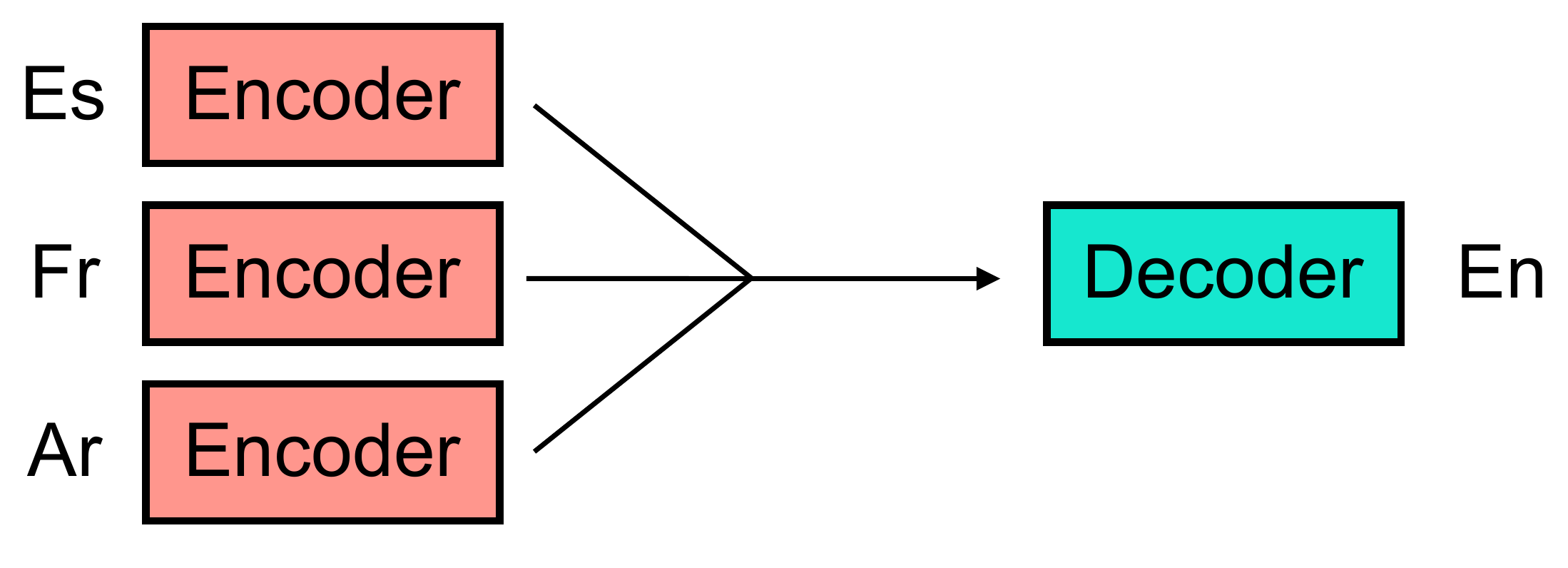}
    \caption{Multi-encoder NMT}
    \label{fig:multi_src}
\end{figure}

Multi-encoder NMT \cite{zoph-knight:2016:N16-1} is similar to the standard attentional NMT framework \cite{bahdanau-cho-bengio:2015:iclr} but uses multiple encoders corresponding to the source languages and a single decoder, as shown in Figure~\ref{fig:multi_src}.
Suppose we have two LSTM-based encoders and
their hidden states and cell states at the end of the inputs are $h_1$, $h_2$ and $c_1$, $c_2$, respectively.
The multi-encoder NMT method initializes its decoder hidden states $h$ and cell state $c$ as follows:
\begin{equation}
h=\tanh (W_c[h_1;h_2]) 
\end{equation}
\begin{equation}
c = c_1 + c_2
\end{equation}

Attention is then defined over each encoder at each time step $t$
and resulting context vectors $c_t^1$ and $c_t^2$, which are concatenated
together with the corresponding decoder hidden state $h_t$
to calculate the final context vector $\tilde{h_t}$. 

\begin{equation}
\tilde{h_t}=\tanh (W_c[h_t;c_t^1;c_t^2])
\end{equation}

The method we base our work upon is largely similar to \newcite{zoph-knight:2016:N16-1}, with the exception of a few details.
Most notably, they used \textit{local-p} attention, which focuses only on a small subset of the source positions for each target word \cite{luong-pham-manning:2015:EMNLP}.
In this work, we used \textit{global} attention, which attends to all words on the source side for each target word, as this is the standard method used in the great majority of recent NMT work. 

\subsection{Mixture of NMT Experts}
\begin{figure}[t]
	\includegraphics[width=7.5cm]{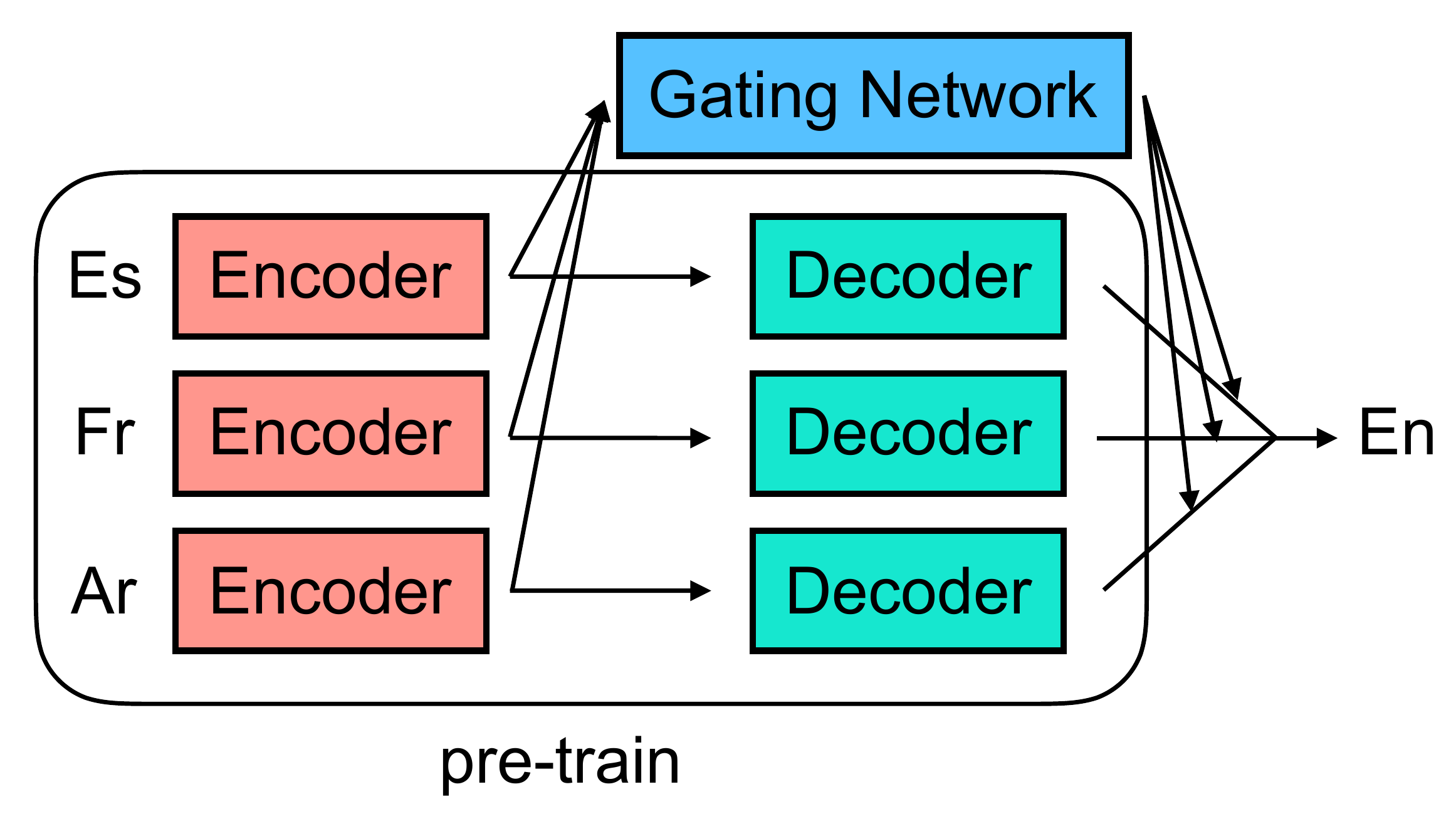}
    \caption{Mixture of NMT Experts}
    \label{fig:moe}
\end{figure}

\newcite{garmash-monz:2016:COLING} proposed another approach to multi-source NMT called \textit{mixture of NMT experts}.
This method ensembles together independently-trained encoder-decoder networks.
Each NMT model is trained using a bilingual corpus with one source language and the target language,
and the outputs from the one-to-one models are summed together, weighted
according to a gating network to control contributions of the probabilities from each model,
as shown in Figure~\ref{fig:moe}.

The mixture of NMT experts determines an output symbol at each time step $t$
from the final output vector $p_t^\epsilon$,
which is the weighted sum of the probability vectors from one-to-one models denoted as follows:

\begin{equation}
p_t^\epsilon = \sum_{j=1}^m g_t^j p_t^j
\end{equation}
where $p_t^j$ and $g_t^j$ are the probability vector from $j$-th model and the corresponding weight at time step $t$, respectively.
$m$ is the number of one-to-one models.
$g_t$ is calculated by the gating network as follows:

\footnotesize
\begin{equation}
g_t=\rm{softmax}(\it{W_{gate}}\tanh(\it{W_{hid}}[f_t^1(x);...f_t^m(x)]))
\end{equation}
\normalsize
where $f_t^j(x)$ is the input vector to the decoder of the $j$-th model,
typically the embedding vector for the output symbol at the previous time step $t$-1.

\section{Multi-Source NMT with Missing Data}
In this work, we examine methods to use incomplete multilingual corpora to improve NMT in a specific language pair.
This allows multi-source techniques to be applied, reaping the benefits of other additional languages even if some translations in these additional languages are missing.
Specifically, we attempt to extend the methods in the previous section to use an incomplete multilingual corpus in this work.

\subsection{Multi-encoder NMT}
\begin{figure}[t]
	\includegraphics[width=7.5cm]{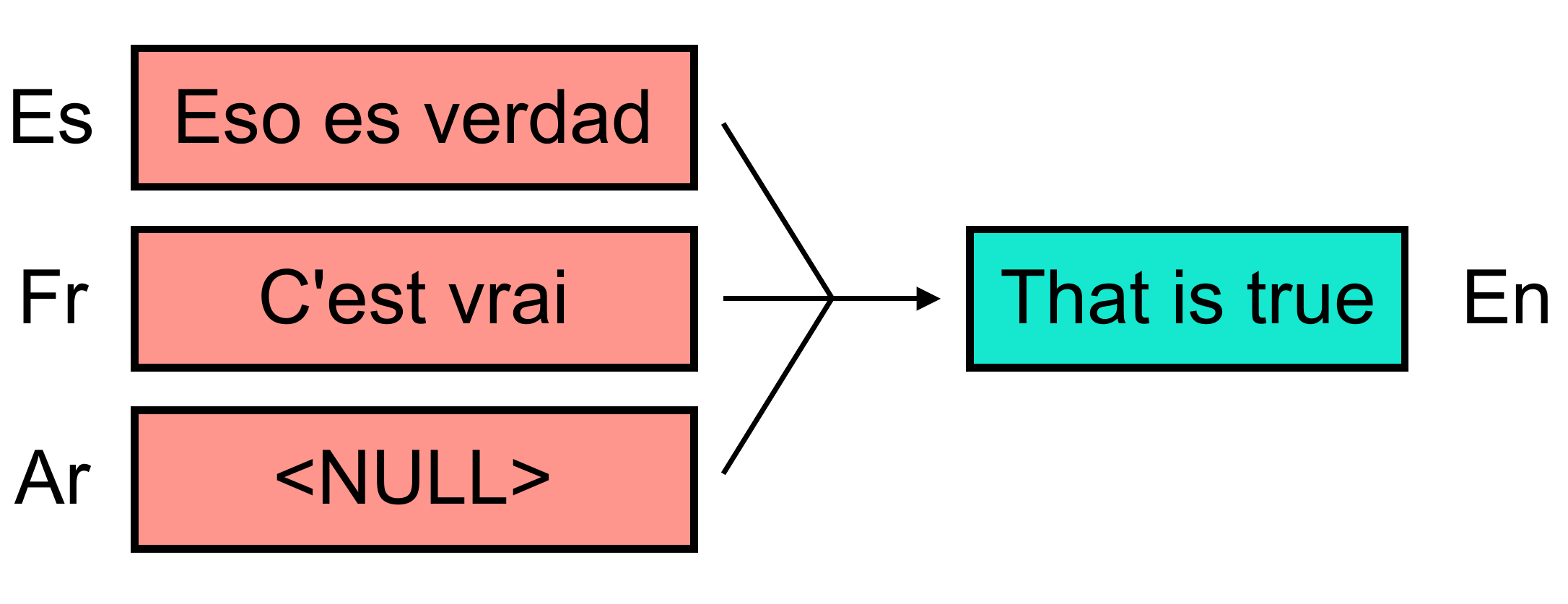}
    \caption{Multi-encoder NMT with a missing input sentence} %\gn{Change to $<$NULL$>$}}
    \label{fig:lack_example}
\end{figure}

In multi-encoder NMT,
each encoder must be provide with an input sentence,
so incomplete multilingual corpora cannot be used as-is.

In this work, we employ a
very simple modification that helps resolve this issue:
replacing each missing input sentence with a special symbol $<$NULL$>$.
The special symbol $<$NULL$>$ can be expected to be basically ignored in multi-encoder NMT,
with the decoder choosing word hypotheses using other input sentences.
Note that this method can be applied easily
to any existing implementation of the multi-encoder NMT with no modification of the codes.

Figure~\ref{fig:lack_example} illustrates the modified multi-encoder NMT method.
Here the source languages are Spanish, French, and Arabic
and the target language is English,
and the Arabic input sentence is missing.
Here, the Spanish and French input sentences are passed into the corresponding encoders
and $<$NULL$>$ is input to the Arabic encoder.

\subsection{Mixture of NMT Experts}
In the mixture of NMT experts method, 
each one-to-one NMT model can be trained independently using incomplete multilingual corpora.
However,
we still need a complete multilingual corpus to train the gating network.

We also employ a special symbol $<$NULL$>$ in the mixture of NMT experts to deal with missing input sentences
in the same way as the multi-encoder NMT described above.
The gating network can also be expected to learn to ignore the outputs from the missing inputs.

\section{Experiments}
We conducted two experiments with different incomplete multilingual corpora.
One is an experiment with a pseudo-incomplete multilingual corpus,
the other is an experiment with an actual incomplete multilingual corpus.   

\subsection{NMT settings}
We describe the settings of common parts for all NMT models: multi-encoder NMT, mixture of NMT experts, and one-to-one NMT.
We used global attention and attention feeding \cite{luong-pham-manning:2015:EMNLP} for the NMT models
and used a bidirectional encoder \cite{bahdanau-cho-bengio:2015:iclr} in their encoders.
The number of units was 512 for the hidden and embedding layers.
Vocabulary size was the most frequent 30,000 words 
in the training data for each source and target languages.
The parameter optimization algorithm was Adam \cite{kingma-ba:2015:iclr} and gradient clipping was set to 5.
The number of hidden state units in the gating network for the mixture of NMT experts experiments was 256.
We used BLEU \cite{papineni-etc:2002:ACL} as the evaluation metric.
We performed early stopping, saving parameter values that had the smallest log perplexities on the validation data and used them when decoding test data.

\subsection{Pseudo-incomplete multilingual corpus (UN6WAY)}
First,
we conducted experiments using a \textit{complete} multilingual corpus
and a \textit{pseudo-incomplete} corpus derived by excluding some sentences from the complete corpus,
to compare the performance in complete and incomplete situations.

\subsubsection{Data}
We used UN6WAY \cite{ZIEMSKI16.1195} as the complete multilingual corpus.
We chose Spanish (Es), French (Fr), and Arabic (Ar) as source languages and English (En) as a target language
The training data in the experiments were the one million sentences from the UN6WAY corpus
whose sentence lengths were less than or equal to 40 words.
We excluded 200,000 sentences for each language for the pseudo-incomplete multilingual corpus
as shown in Table~\ref{table:pseudo}.
``Sentence number'' in Table 1 represents the line number in the corpus,
and the {\em x} means the part removed for the incomplete multilingual corpus.
We also chose 1,000 and 4,000 sentences for validation and test from the UN6WAY corpus,
apart from the training data.
Note that the validation and test data here had no missing translations.

\subsubsection{Setup}
\begin{table}[t]
	\center
    \begin{tabular}{|l||c|c|c|c|} \hline
    	Sentence No. & Es & Fr & Ar & En \\ \hline
        1-200,000         & x &   &   &  \\ \hline
        200,001-400,000   &   &   & x &  \\ \hline
        400,001-600,000   &   & x &   &  \\ \hline
        600,001-800,000   &   &   &   &  \\ \hline
    \end{tabular}
    \caption{Settings of the pseudo-incomplete UN multilingual corpus (x means that this part was deleted)}
    \label{table:pseudo}
\end{table}

We compared multi-encoder NMT and the mixture of NMT experts in the complete and incomplete situations.
The three one-to-one NMT systems, Es-En, Fr-En, and Ar-En, which were used as sub-models in the mixture of NMT experts,
were also compared for reference.

First, we conducted experiments using all of the one million sentences in the complete multilingual corpus, \textit{Complete (0.8M)}.
In case of the mixture of NMT experts,
the gating network was trained using the one million sentences.

Then, we tested in the incomplete data situation.
Here there were just 200,000 complete multilingual sentences (sentence No. 600,001-800,000), \textit{Complete (0.2M)}.
Here, a standard
multi-encoder NMT and mixture of NMT experts could be trained using this complete data.
On the other hand, the multi-source NMT with $<$NULL$>$
could be trained using 800,000 sentences (sentence No. 1-800,000), \textit{Pseudo-incomplete (0.8M)}.
Each one-to-one NMT could be trained using these 800,000 sentences,
but the missing sentences replaced with the $<$NULL$>$ tokens were excluded
so resulting 600,000 sentences were actually used.

\subsubsection{Results}

\begin{table*}[t]
    \hspace*{-2.5em}
	\begin{center}
    \begin{tabular}{|l||c|c|c|c|c|} \hline
    	\multirow{2}{*}{Condition}        & \multicolumn{3}{|c|}{One-to-one}     & \multirow{2}{*}{Multi-encoder} & \multirow{2}{*}{Mix. NMT Experts} \\ \cline{2-4}
                                          & Es-En & Fr-En & Ar-En                & & \\ \hline
        Complete (0.8M)                     & 31.87 & 25.78 & 23.08                & 37.55 (+5.68)$^*$  & 33.28 (+1.41) \\ \hline
        Complete (0.2M)                   & 27.62 & 22.01 & 17.88                & 31.24 (+3.62)  & 32.16 (+4.54) \\ \hline
        \textit{Pseudo-incomplete} (0.8M) & 30.98 & 25.62 & 22.02                & \textbf{36.43} (+5.45)$^*$ & 32.44 (+1.47) \\ \hline
    \end{tabular}
    \caption{Results in BLEU for one-to-one and multi-source (\{Es, Fr, Ar\}-to-En) translation on UN6WAY data (parentheses are BLEU gains against the best one-to-one results). $^*$ indicates the difference from mixture of NMT experts is statistically significant ($p < 0.01$).}
    
    \label{table:un6way}
    \end{center}
\end{table*}

\mbox{Table~\ref{table:un6way}} shows the results in BLEU.
The multi-source approaches achieved consistent improvements over the one-to-one NMTs in the all conditions,
as demonstrated in previous multi-source NMT studies.
Our main focus here is Pseudo-incomplete (0.8M),
in which the multi-source results were slightly worse than those in Complete (0.8M) but better than those in Complete (0.2M).
This suggests the additional use of incomplete corpora is beneficial in multi-source NMT
compared to the use of only the complete parts of the corpus,
even if just through the simple modification of replacing missing sentences with $<$NULL$>$.

With respect to the difference between the multi-encoder NMT and mixture of NMT experts,
the multi-encoder achieved much higher BLEU in Pseudo-incomplete (0.8M) and Complete (1M), but this was not the case in Complete (0.2M).
One possible reason here is the model complexity;
the multi-encoder NMT uses a large single model while one-to-one sub-models in the mixture of NMT experts can be trained independently.

\begin{table}[t]
	\center
    \begin{tabular}{|l||c|c|c|} \hline
    	Source & Training & Valid. & Test \\ \hline\hline
        \multicolumn{4}{|l|}{\{En, Fr, Pt (br)\}-to-Es} \\ \hline
        English    & 189,062  & 4,076 & 5,451  \\ \hline
        French     & 170,607  & 3,719 & 4,686  \\ \hline
        Portuguese (br)& 166,205  & 3,623 & 4,647  \\ \hline
        \hline
        \multicolumn{4}{|l|}{\{En, Es, Pt (br)\}-to-Fr} \\ \hline
        English     & 185,405 & 4,164 & 4,753  \\ \hline
        Spanish     & 170,607 & 3,719 & 4,686  \\ \hline
        Portuguese (br) & 164,630 & 3,668 & 4,289  \\ \hline
        \hline
        \multicolumn{4}{|l|}{\{En, Es, Fr\}-to-Pt (br)} \\ \hline
        English    & 177,895 & 3,880 & 4,742  \\ \hline
        Spanish    & 166,205 & 3,623 & 4,647  \\ \hline
        French     & 164,630 & 3,668 & 4,289  \\ \hline     
    \end{tabular}
    \caption{Data statistics in the tasks on TED data (in the number of sentences). Note that the number of target sentences is equal to that of English for each task.}
    \label{table:wit3_data}
\end{table}

\begin{table}[t]
	\center
    \begin{tabular}{|l||c|c|c|} \hline
    	Target & Training & Valid. & Test \\ \hline\hline
        Spanish & 83.4 & 85.0 & 78.2 \\ \hline
        French    & 85.0 & 83.2 & 89.7  \\ \hline
        Portuguese (br)     & 88.6 & 89.3 & 90.0  \\ \hline     
    \end{tabular}
    \caption{The percentage of data without missing sentences on TED data.}
    \label{table:wit3_complete_percentage}
\end{table}

\subsection{An actual incomplete multilingual corpus (TED Talks)}
\subsubsection{Data}
We used a collection of transcriptions of TED Talks and their multilingual translations.
Because these translations are created by volunteers,
and the number of translations for each language is dependent on the number of volunteers who created them,
this collection is an actual incomplete multilingual corpus.
The great majority of the talks are basically in English, so we chose English as a source language.
We used three translations in other languages for our multi-source scenario: Spanish, French, Brazilian Portuguese.
We prepared three tasks choosing one of each of these three languages as the target language and the others as the additional source languages.
\mbox{Table~\ref{table:wit3_data}} shows
the number of available sentences in these tasks,
chosen so that their lengths are less than or equal to 40 words.

\subsubsection{Setup}
We compared multi-encoder NMT, mixture of NMT experts and one-to-one NMT with English as the source language.
The validation and test data for these experiments were also incomplete.
This is in contrast to the experiments on UN6WAY where the test and validation data were complete, and thus this setting is arguable of more practical use.

\subsubsection{Results}
\mbox{Table~\ref{table:wit3_result}} shows the results in BLEU and BLEU gains with respect to the one-to-one results.
All the differences are statistically significant ($p < 0.01$) by significance tests with bootstrap resampling \cite{koehn:2004:EMNLP}.
The multi-source NMTs achieved consistent improvements over the one-to-one baseline as expected,
but the BLEU gains were smaller than those in the previous experiments using the UN6WAY data.
This is possibly because the baseline performance was relatively low compared with the previous experiments and the size of available resources was also smaller.

In comparison between the multi-source NMT and the mixture of NMT experts, results were mixed;
the mixture of NMT experts was better in the task to French.

\subsubsection{Discussion}
\begin{figure}[h!] 
	\subfigure[TED: \{En,Fr,Pt (br)\}-to-Es]{
      \includegraphics[width=8.0cm]{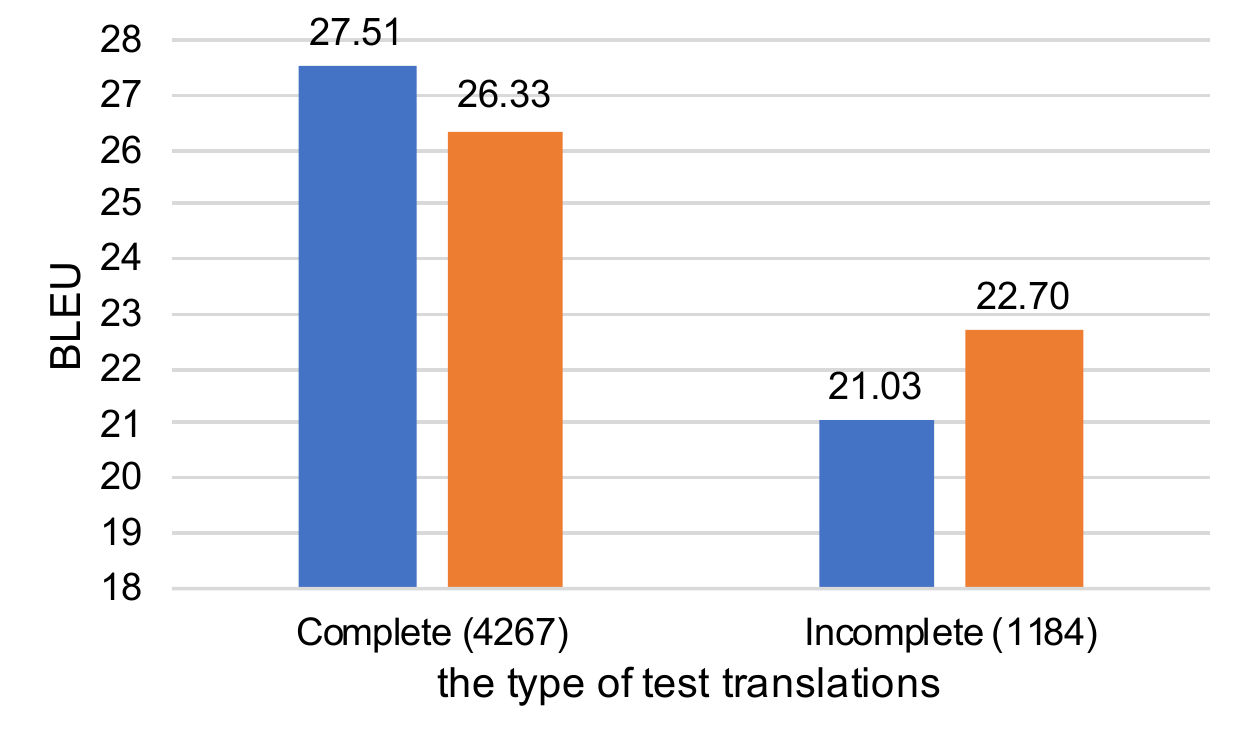}}
	\subfigure[TED: \{En,Es,Pt (br)\}-to-Fr]{
      \includegraphics[width=8.0cm]{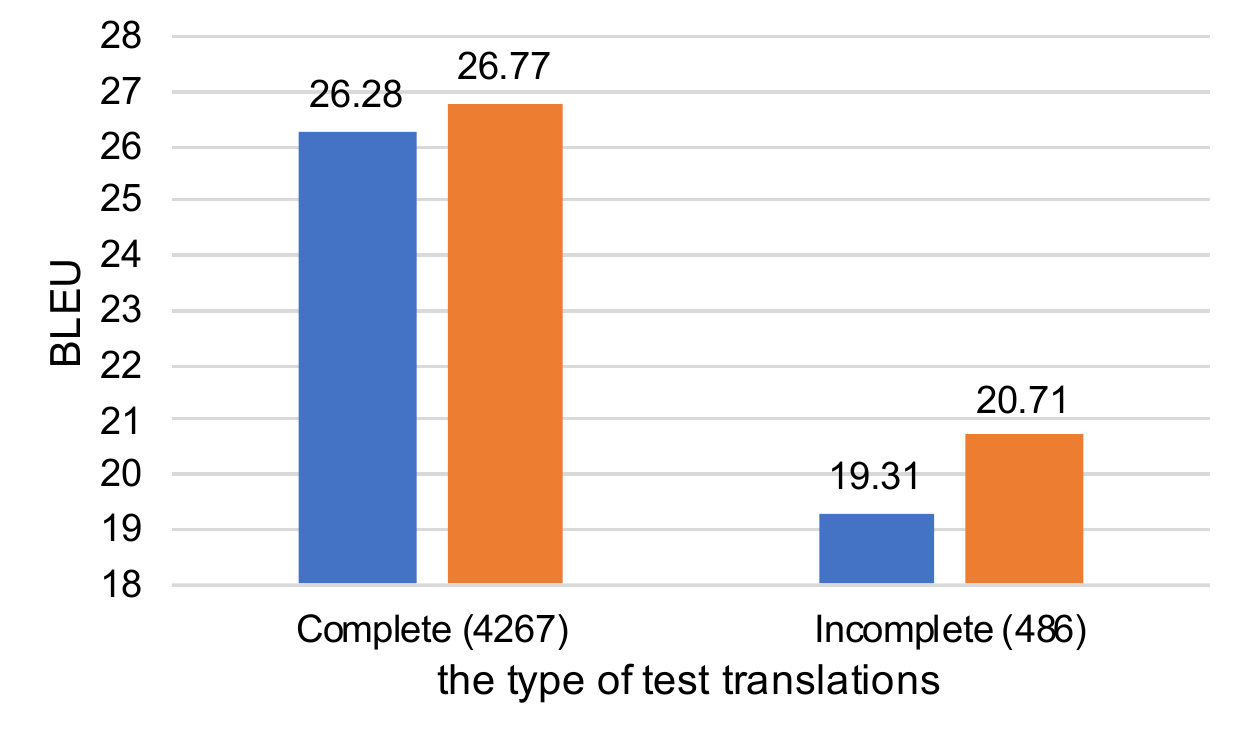}}
	\subfigure[TED: \{En,Es,Fr\}-to-Pt (br)]{
      \includegraphics[width=8.0cm]{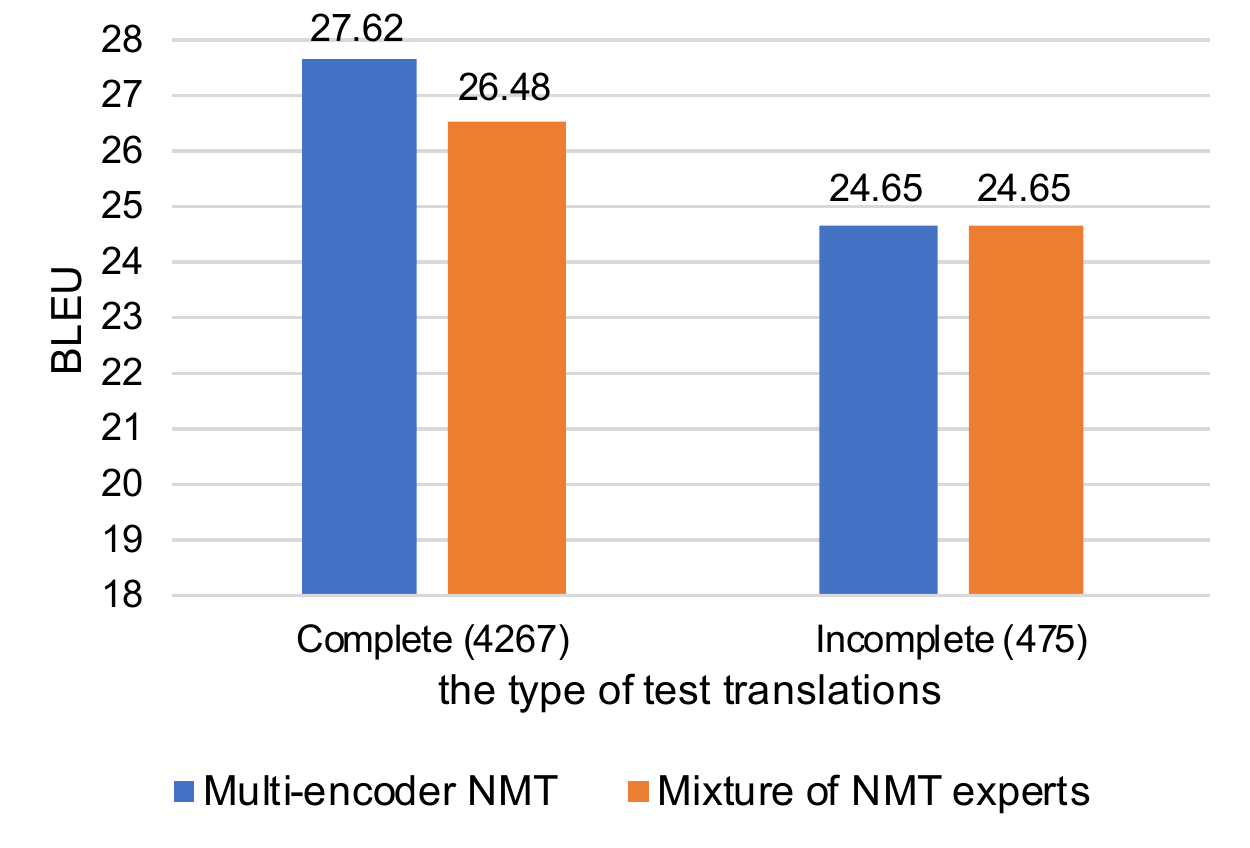}}
    \caption{Detailed comparison of BLEU in TED test data. {\em Complete} means the part of test data, in which there is no missing translation, and {\em incomplete} means that, in which there are some missing translation. The number in a parenthesis is the number of translations.}
    \label{fig:ted_details}
\end{figure}

We analyzed the results using the TED data in detail to investigate the mixed results above.
Figure~\ref{fig:ted_details} (in the last page) shows the breakdown of BLEU in the test data,
separating the results for complete and incomplete multilingual inputs.
When all source sentences are present in the test data,
multi-encoder NMT has better performance than mixture of NMT experts except for \{En, Es, Pt (br)\}-to-Fr. 
However, when the input is incomplete, mixture of NMT experts achieves performance better than or equal to multi-encoder NMT.
From this result, we can assume that mixture of NMT experts, with its explicit gating network,
is better at ignoring the irrelevant missing sentences.
It's possible that if we designed a better attention strategy for multi-encoder NMT we may be able to resolve this problem.
These analyses would support the results using the pseudo incomplete data shown in Table~\ref{table:un6way},
where the validation and test data were complete.

\begin{table*}[t]
	\center
    \begin{tabular}{|l||c|c|c|} \hline
    	\multirow{2}{*}{Task} & One-to-one & \multirow{2}{*}{Multi-encoder} & \multirow{2}{*}{Mix. NMT Experts}  \\
             & (En-to-target) & & \\ \hline\hline
        \{En, Fr, Pt (br)\}-to-Es & 24.32 & \textbf{26.01} (+1.69) &         25.51  (+1.19) \\ \hline
        \{En, Es, Pt (br)\}-to-Fr & 24.54 &         25.62  (+1.08) & \textbf{26.23} (+1.69)  \\ \hline
        \{En, Es, Fr\}-to-Pt (br) & 25.14 & \textbf{27.36} (+2.22) &         26.39  (+1.25)  \\ \hline
    \end{tabular}
     \caption{
     Results in BLEU (and BLEU gains) by one-to-one and multi-source NMT on TED data. Note that the target language in each row differs so the results in different rows cannot be compared directly. All the differences are statistically significant ($p < 0.01$).} 
     \label{table:wit3_result}
\end{table*}

 \subsubsection{Translation examples}

 \begin{table*}[h]
  \begin{center}
   \begin{tabular}{|l||l|r|}
    \hline
    Type             & Sentence & BLEU+1 \\ \hline\hline

    \multicolumn{3}{|l|}{Example (1)} \\ \hline
    Source (En)      &Then I started looking at the business model. & \\
    Source (Fr)      & Puis j'ai regardé le modèle économique. & \\
    Source (Pt (br)) & $<$NULL$>$  & \\ \hline
    Reference        & Después empecé a ver el modelo de negocio. & \\ \hline
    En-to-Es         &  Luego empecé a estudiar el modelo empresarial. & 0.266\\
    Multi-encoder    & Luego empecé a mirar el modelo empresarial. & 0.266\\
    Mix. NMT experts &  Luego empecé a ver el modelo de negocios. & 0.726\\ \hline\hline
   
    \multicolumn{3}{|l|}{Example (2)} \\ \hline
    Source (En)      & Sometimes they agree.& \\
    Source (Fr)      & $<$NULL$>$ & \\
    Source (Pt (br)) & $<$NULL$>$ & \\ \hline
    Reference        & A veces están de acuerdo. & \\ \hline
    En-to-Es         &   A veces están de acuerdo. & 1.000\\
    Multi-encoder    &  A veces están de acuerdo. & 1.000\\
    Mix. NMT experts &  A veces están de acuerdo.& 1.000\\ \hline
   \end{tabular}
   \caption{Translation examples in \{English, French, Brazilian Portuguese\}-to-Spanish translation.}
   \label{table:examples}
  \end{center}
 \end{table*}

 Table~\ref{table:examples} shows a couple of translation examples in the \{English, French, Brazilian Portuguese\}-to-Spanish experiment.
 In Example(1), 
 BLEU+1 of mixture of NMT Experts is larger than one-to-one (English-to-Spanish) because of the French sentence, although the source sentence of Brazilian Portuguese is missing.
BLEU+1 of multi-encoder is same as one-to-one, but the generation word is different.
The word of "minar" is generated from multi-encoder, and "estudiar" is generated from one-to-one.
"minar" means "look" in English, and "estudiar" means "study", so the meaning of sentence which was generated from multi-encoder is close to the reference one than that from one-to-one.
Besides the word of "ver" which is generated from mixture of NMT experts  meas "see" in English, so the sentence of multi-encoder  is more appropriate than the reference sentence.

In Example(2), there is only the English sentence in the source sentences.
We can see that sentences which are generated from all models are same as the reference sentences, although French and Brazilian Portuguese sentences are missing.
Therefore multi-source NMT models  work properly even if there are missing sentences.

\section{Related Work}
In this paper, we examined strategies for multi-source NMT.
On the other hand, there are other strategies for multilingual NMT
that do not use multiple source sentences as their input.
\citet{dong-etc:2015:ACL} proposed a method for multi-target NMT.
Their method is using one sharing encoder and decoders corresponding to the number of target languages. 
\citet{firat-cho-bengio:2016:N16-1} proposed a method for multi-source multi-target NMT
using multiple encoders and decoders with a shared attention mechanism.
\citet{johonson-etc:2016:CoRR} and \citet{ha2016toward} proposed multi-source and multi-target NMT
using one encoder and one decoder, and sharing all parameters with all languages. 
Notably, these methods use multilingual data to better train one-to-one NMT systems.
However, our motivation of this study is to improve NMT further
by the help of other translations that are available on the source side at test time, 
and thus their approaches are different from ours.

\section{Conclusion}
In this paper, we examined approaches for multi-source NMT
using \textit{incomplete} multilingual corpus
in which each missing input sentences is replaced by a special symbol $<$NULL$>$.
The experimental results with simulated and actual incomplete multilingual corpora show that
this simple modification allows us to effectively use all available translations at both training and test time.

The performance of multi-source NMT depends on source and target languages, and the size of missing data.
As future work, we will investigate the relation of the languages included in the multiple sources
and the number of missing inputs to the translation accuracy in multi-source scenarios.

\section{Acknowledgement}
Part of this work was supported by JSPS KAKENHI Grant Numbers and JP16H05873
and JP17H06101.

\bibliographystyle{acl_natbib}
%\bibliography{wnmt2018}

\begin{thebibliography}{16}
\expandafter\ifx\csname natexlab\endcsname\relax\def\natexlab#1{#1}\fi

\bibitem[{Bahdanau et~al.(2015)Bahdanau, Cho, and
  Bengio}]{bahdanau-cho-bengio:2015:iclr}
Dzmitry Bahdanau, Kyunghyun Cho, and Yoshua Bengio. 2015.
\newblock \href {http://arxiv.org/abs/1409.0473} {{Neural Machine Translation
  by Jointly Learning to Align and Translate}}.
\newblock In \emph{Proceedings of the 3rd International Conference on Learning
  Representations}.

\bibitem[{Cettolo et~al.(2012)Cettolo, Girardi, and
  Federico}]{Cettolo-etc:2012:eamt}
Mauro Cettolo, Christian Girardi, and Marcello Federico. 2012.
\newblock \href {https://wit3.fbk.eu/papers/WIT3-EAMT2012.pdf} {{WIT$^{3}$: Web
  Inventory of Transcribed and Translated Talks}}.
\newblock In \emph{Proceedings of the 16th EAMT Conference}, pages 261--268.

\bibitem[{Dong et~al.(2015)Dong, Wu, He, Yu, and Wang}]{dong-etc:2015:ACL}
Daxiang Dong, Hua Wu, Wei He, Dianhai Yu, and Haifeng Wang. 2015.
\newblock \href {http://www.aclweb.org/anthology/P15-1166} {{Multi-Task
  Learning for Multiple Language Translation}}.
\newblock In \emph{Proceedings of the 53rd Annual Meeting of the Association
  for Computational Linguistics}, pages 1723--1732, Beijing, China. Association
  for Computational Linguistics.

\bibitem[{Firat et~al.(2016)Firat, Cho, and
  Bengio}]{firat-cho-bengio:2016:N16-1}
Orhan Firat, Kyunghyun Cho, and Yoshua Bengio. 2016.
\newblock \href {http://www.aclweb.org/anthology/N16-1101} {{Multi-Way,
  Multilingual Neural Machine Translation with a Shared Attention Mechanism}}.
\newblock In \emph{Proceedings of the 2016 Conference of the North American
  Chapter of the Association for Computational Linguistics: Human Language
  Technologies}, pages 866--875, San Diego, California. Association for
  Computational Linguistics.

\bibitem[{Garmash and Monz(2016)}]{garmash-monz:2016:COLING}
Ekaterina Garmash and Christof Monz. 2016.
\newblock \href {http://aclweb.org/anthology/C16-1133} {{Ensemble Learning for
  Multi-Source Neural Machine Translation}}.
\newblock In \emph{Proceedings of COLING 2016, the 26th International
  Conference on Computational Linguistics: Technical Papers}, pages 1409--1418,
  Osaka, Japan. The COLING 2016 Organizing Committee.

\bibitem[{Ha et~al.(2016)Ha, Niehues, and Waibel}]{ha2016toward}
Thanh-Le Ha, Jan Niehues, and Alexander Waibel. 2016.
\newblock \href
  {https://workshop2016.iwslt.org/downloads/IWSLT_2016_paper_5.pdf} {{Toward
  Multilingual Neural Machine Translation with Universal Encoder and Decoder}}.
\newblock In \emph{Proceedings of the 13th International Workshop on Spoken
  Language Translation}, Seattle, Washington.

\bibitem[{Johonson et~al.(2017)Johonson, Schuster, V.~Le, Krikun, Wu, Chen,
  Thorat, Viégas, Wattenberg, Corrado, Hughes, and
  Dean}]{johonson-etc:2016:CoRR}
Melvin Johonson, Mike Schuster, Quoc V.~Le, Maxim Krikun, Yonghui Wu, Zhifeng
  Chen, Nikhil Thorat, Fernanda Viégas, Martin Wattenberg, Greg Corrado,
  Macduff Hughes, and Jeffrey Dean. 2017.
\newblock \href {https://www.aclweb.org/anthology/Q/Q17/Q17-1024.pdf}
  {{Google's Multilingual Neural Machine Translation System: Enabling Zero-Shot
  Translation}}.
\newblock \emph{Transactions of the Association for Computational Linguistics,
  vol. 5}, pages 339--351.

\bibitem[{Kingma and Ba(2015)}]{kingma-ba:2015:iclr}
Diederik P.~Kingma Kingma and Jimmy Ba. 2015.
\newblock \href {https://arxiv.org/abs/1412.6980} {{Adam: A Method for
  Stochastic Optimization}}.
\newblock In \emph{Proceedings of the 3rd International Conference on Learning
  Representations}.

\bibitem[{Koehn(2004)}]{koehn:2004:EMNLP}
Philipp Koehn. 2004.
\newblock {Statistical Significance Tests for Machine Translation Evaluation}.
\newblock In \emph{Proceedings of EMNLP 2004}, pages 388--395, Barcelona,
  Spain. Association for Computational Linguistics.

\bibitem[{Koehn(2005)}]{koehn2005epc}
Philipp Koehn. 2005.
\newblock \href {http://mt-archive.info/MTS-2005-Koehn.pdf} {{Europarl: A
  Parallel Corpus for Statistical Machine Translation}}.
\newblock In \emph{{Conference Proceedings: the tenth Machine Translation
  Summit}}, pages 79--86, Phuket, Thailand. AAMT, AAMT.

\bibitem[{Luong et~al.(2015)Luong, Pham, and
  Manning}]{luong-pham-manning:2015:EMNLP}
Thang Luong, Hieu Pham, and Christopher~D. Manning. 2015.
\newblock \href {http://aclweb.org/anthology/D15-1166} {{Effective Approaches
  to Attention-based Neural Machine Translation}}.
\newblock In \emph{Proceedings of the 2015 Conference on Empirical Methods in
  Natural Language Processing}, pages 1412--1421, Lisbon, Portugal. Association
  for Computational Linguistics.

\bibitem[{Och and Ney(2001)}]{Och2001}
Franz~Josef Och and Hermann Ney. 2001.
\newblock {Statistical Multi-Source Translation}.
\newblock In \emph{Proceedings of the eighth Machine Translation Summit (MT
  Summit VIII)}, pages 253--258.

\bibitem[{Papineni et~al.(2002)Papineni, Roukos, Ward, and
  Zhu}]{papineni-etc:2002:ACL}
Kishore Papineni, Salim Roukos, Todd Ward, and Wei-Jing Zhu. 2002.
\newblock {{\sc BLEU}: a Method for Automatic Evaluation of Machine
  Translation}.
\newblock In \emph{Proceedings of the 40th Annual Meeting of the Association
  for Computational Linguistics (ACL)}, pages 311--318, Philadelphia.

\bibitem[{Tiedemann(2009)}]{Tiedemann:RANLP5}
J\"org Tiedemann. 2009.
\newblock {News from OPUS - A Collection of Multilingual Parallel Corpora with
  Tools and Interfaces}.
\newblock In N.~Nicolov, K.~Bontcheva, G.~Angelova, and R.~Mitkov, editors,
  \emph{Recent Advances in Natural Language Processing}, volume~V, pages
  237--248. John Benjamins, Amsterdam/Philadelphia, Borovets, Bulgaria.

\bibitem[{Ziemski et~al.(2016)Ziemski, Junczys-Dowmunt, and
  Pouliquen}]{ZIEMSKI16.1195}
Michał Ziemski, Marcin Junczys-Dowmunt, and Bruno Pouliquen. 2016.
\newblock {The United Nations Parallel Corpus v1.0}.
\newblock In \emph{Proceedings of the Tenth International Conference on
  Language Resources and Evaluation (LREC 2016)}, Paris, France. European
  Language Resources Association (ELRA).

\bibitem[{Zoph and Knight(2016)}]{zoph-knight:2016:N16-1}
Barret Zoph and Kevin Knight. 2016.
\newblock \href {http://www.aclweb.org/anthology/N16-1004} {{Multi-Source
  Neural Translation}}.
\newblock In \emph{Proceedings of the 2016 Conference of the North American
  Chapter of the Association for Computational Linguistics: Human Language
  Technologies}, pages 30--34, San Diego, California. Association for
  Computational Linguistics.

\end{thebibliography}

\end{document}